\documentclass[11pt]{article}

\usepackage[preprint]{acl}

\usepackage{times}
\usepackage{latexsym}

\usepackage[T1]{fontenc}

\usepackage[utf8]{inputenc}

\usepackage{microtype}

\usepackage{inconsolata}

\usepackage{graphicx}
\usepackage{booktabs}

\title{Neural Sabermetrics with World Model:\\
Play-by-play Predictive Modeling with Large Language Model}

\author{
  Young Jin Ahn \quad Yiyang Du \quad Zheyuan Zhang \quad Haisen Kang \\
  Carnegie Mellon University \\
  \texttt{\{youngjia, yiyangd, zheyuan4, haisenk\}@andrew.cmu.edu}
}

\begin{document}
\maketitle
\begin{abstract}
Classical sabermetrics has profoundly shaped baseball analytics by summarizing long histories of play into compact statistics. While these metrics are invaluable for valuation and retrospective analysis, they do not define a generative model of how baseball games unfold pitch by pitch, leaving most existing approaches limited to single-step prediction or post-hoc analysis. 
In this work, we present \emph{Neural Sabermetrics with World Model}, a Large Language Model (LLM) based play-by-play world model for baseball. We cast baseball games as long auto-regressive sequences of events and continuously pretrain a single LLM on more than ten years of Major League Baseball (MLB) tracking data, comprising over seven million pitch sequences and approximately three billion tokens. The resulting model is capable of predicting multiple aspects of game evolution within a unified framework.
We evaluate our model on both in-distribution regular-season data and out-of-distribution postseason games and compare against strong neural baselines from prior work. Despite using a single backbone model, our approach outperforms the performance of existing baselines, (1) correctly predicting approximately 64\% of next pitches within a plate appearance and (2) 78\% of batter swing decisions, suggesting that LLMs can serve as effective world models for sports.
\end{abstract}

\section{Introduction}

Sabermetrics has revolutionized the analysis of baseball by translating complex game histories into interpretable summary statistics. These metrics provide powerful tools for evaluating player performance and team value \cite{b9b2abde784b485588054b154f7a7226, usingml}. However, classical sabermetrics is fundamentally descriptive: it compresses past events into fixed statistics and does not define a generative process capable of simulating or predicting how a game unfolds pitch by pitch.

As a consequence, most traditional and machine-learning-based baseball models focus on narrow prediction tasks, such as pitch type classification or outcome prediction, using carefully engineered features and shallow temporal context \cite{yifanpredict, yulstm, gopal2024baseball}. While these approaches have demonstrated measurable improvements over simple baselines, they typically operate at a single-step level and lack a unified representation of game dynamics. Moreover, they struggle to generalize across contexts with substantial distribution shifts, such as the transition from regular-season to postseason play.

Recent advances in sequence modeling and large language models (LLMs) have enabled a different perspective on structured decision-making problems. In domains such as robotics, reinforcement learning, and control, \emph{world models} aim to learn a generative model of environment dynamics that can predict future states conditioned on past observations and actions \cite{ha2018world, hafner2020dreamer}. In parallel, large language models trained with auto-regressive objectives have demonstrated a remarkable ability to internalize long-range dependencies, latent structure, and domain-specific dynamics when data are represented as sequences of tokens \cite{brown2020language, touvron2023llamaopenefficientfoundation}.

\begin{figure*}[t]
    \centering
    \includegraphics[width=0.8\linewidth]{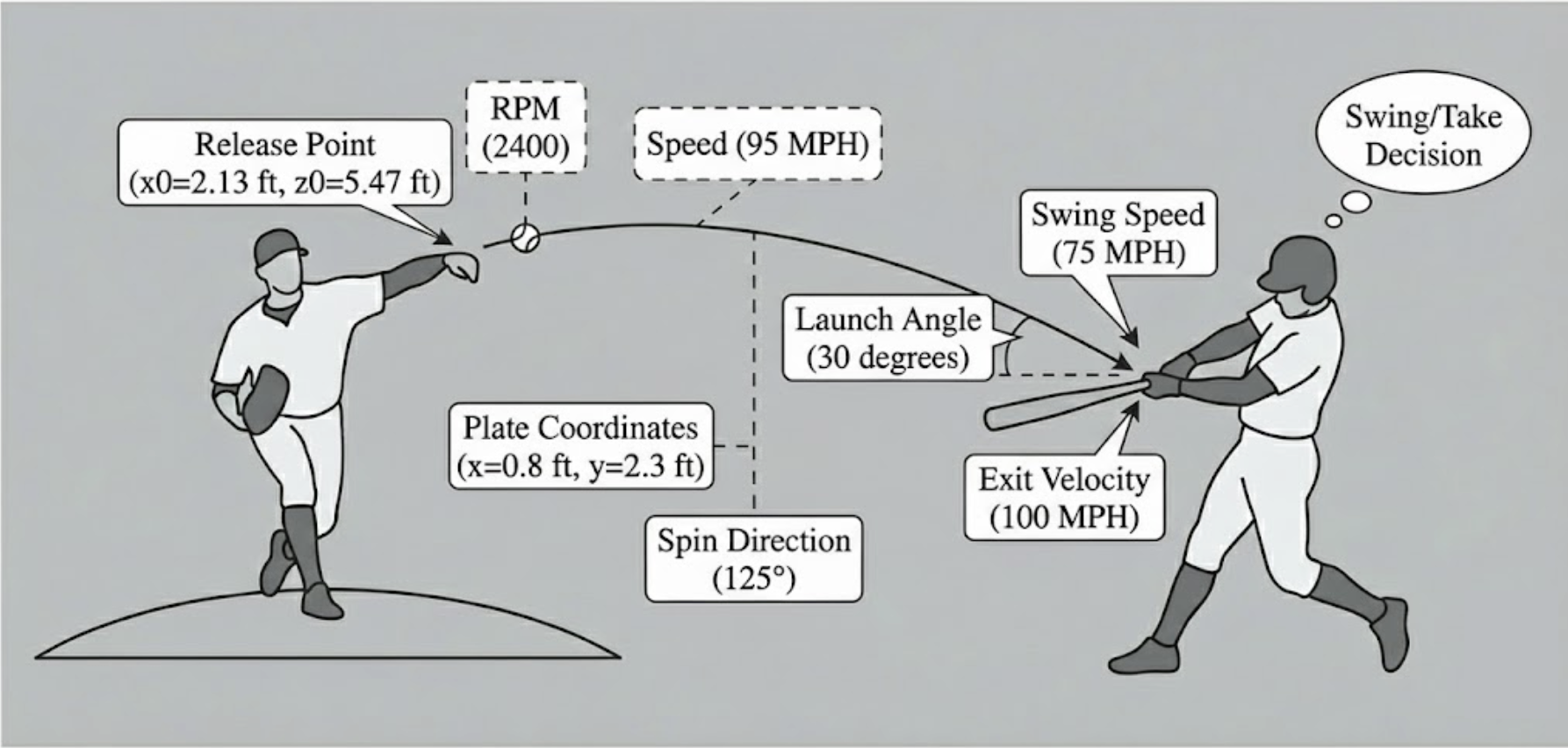}
    \caption{Demonstration of the data. All tracking data from the game event are serialized in sequential texts.}
    \label{fig:data}
\end{figure*}

Inspired by these developments, we explore whether baseball tracking data itself can be treated as a form of language, and whether an LLM can be trained as a \emph{play-by-play world model} for sports. Unlike prior LLM-based work in sports, which primarily focuses on text understanding, summarization, or commentary generation \cite{sportsbert, COOK2024112219, kang2025diamondllmdrivenagentcontextaware}, our goal is to use a large language model directly for next-event prediction in the game itself.

In this paper, we present \emph{Neural Sabermetrics with World Model}, an LLM-based system trained to model the sequential dynamics of baseball at the pitch level. We continuously pretrain a single backbone language model on more than ten years of MLB tracking data, representing over seven million pitch sequences and approximately three billion tokens. Each game is serialized into a long sequence encoding game context, pitch information, batter decisions, and outcomes, enabling the model to learn temporal and strategic dependencies across entire plate appearances and games.

We evaluate our model on multiple downstream prediction tasks, including pitch type prediction and batter swing decision prediction, and compare against strong neural baselines from prior work. Crucially, we train exclusively on regular-season data and evaluate on postseason games to assess robustness under realistic distribution shifts. Our results show that a single LLM backbone can achieve competitive or state-of-the-art performance across tasks, while naturally supporting extensibility to new data modalities and prediction objectives.

Together, our findings suggest that world models offer a promising direction for moving beyond descriptive sabermetrics toward predictive, generative, and simulation-based modeling of sports.

\begin{figure*}[t]
    \centering
    \includegraphics[width=0.8\linewidth]{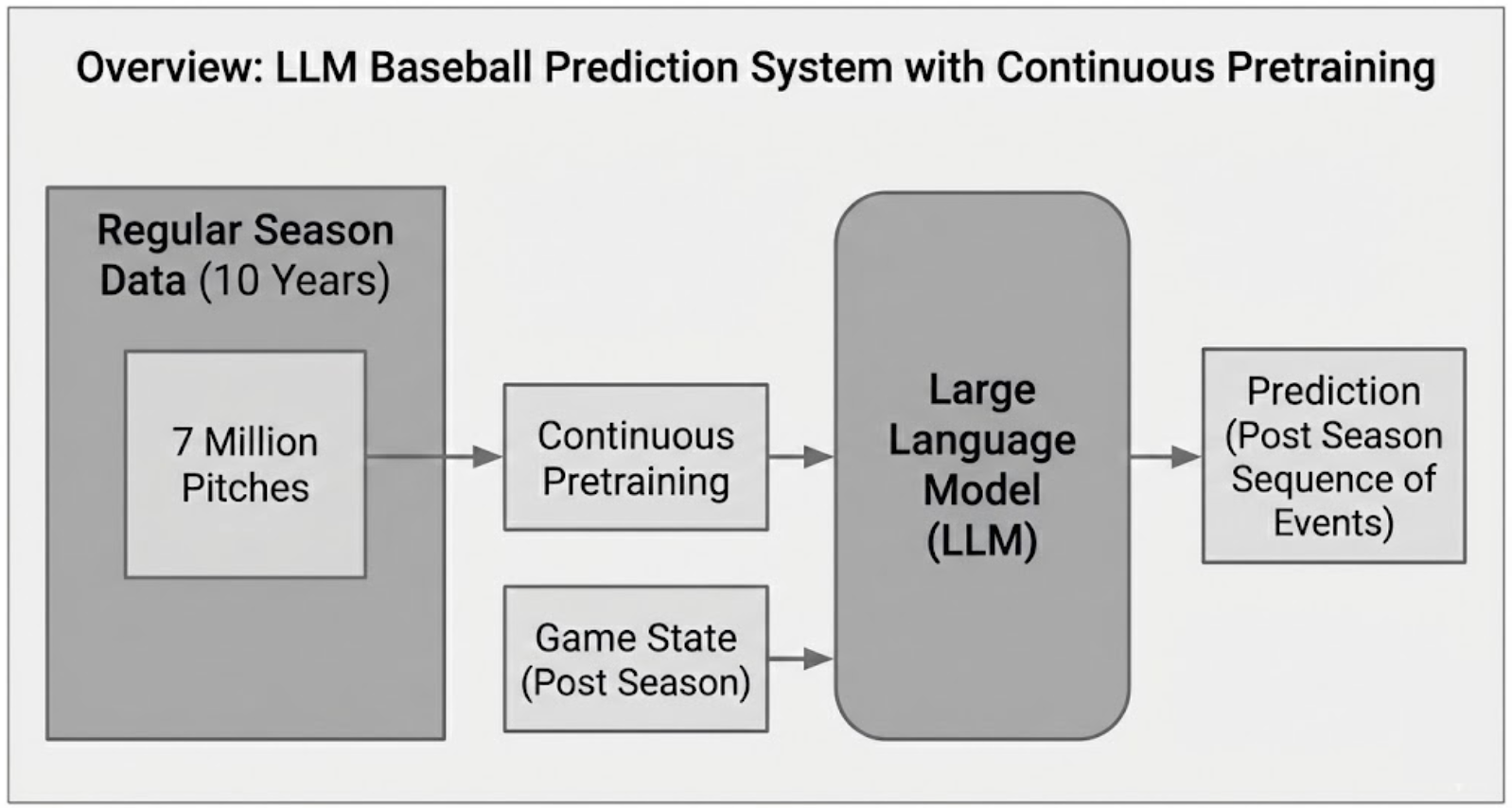}
    \caption{Overview of the training framework. We use 11 years worth of regular season MLB data (3 billion tokens, 7 million pitch sequences) for continuous pretraining and game event predictions.}
    \label{fig:method}
\end{figure*}

\section{Related Work}

\subsection{Sequential Models for Sports Analytics}

Early work on sports prediction has explored sequential models tailored to specific sports. In basketball, \citet{VRACAR201658} proposed a play-by-play simulator conditioned on game context. In soccer, \citet{mendes2024estimating} introduced Large Event Models for next-event prediction. These approaches demonstrate the value of sequence modeling but are highly sport-specific and difficult to transfer to baseball, where pitch-level dynamics dominate.

In baseball, \citet{yifanpredict} introduced a recurrent neural network for binary fastball prediction, showing that temporal information and player embeddings improve over static baselines. \citet{yulstm} extended this line of work with an attention-based LSTM for multi-class pitch type prediction. More recently, \citet{gopal2024baseball} proposed a feed-forward neural network for predicting pitch outcomes and used it as a transition model in at-bat simulations. While effective within narrow scopes, these models are task-specific and lack a unified generative view of the game.

\subsection{LLMs for Sports Understanding}

Large language models have recently been applied to sports-related text and event understanding. SportsBERT \cite{sportsbert} pretrains a masked language model on millions of sports articles, while LLM-Commentator \cite{COOK2024112219} fine-tunes LLaMA for commentary generation. DIAMOND \cite{kang2025diamondllmdrivenagentcontextaware} explores LLM-based highlight summarization in baseball. However, existing LLM-based systems focus primarily on natural language generation or summarization; leveraging LLMs as \emph{world models} for play-by-play prediction in baseball remains largely unexplored.

Our work bridges this gap by framing baseball tracking data itself as a language and training an LLM to model the underlying generative process of pitches and outcomes.

\section{Method}

We model baseball games as long auto-regressive sequences of discrete and continuous events, where each pitch and its outcome are conditioned on all preceding context. Our system follows a continuous pretraining paradigm, using a single large language model as the backbone world model.

\subsection{Event Representation}

Each game is serialized into a textual sequence that includes global game context (teams, venue, weather), inning-level state (outs, score, base runners), and pitch-level information (pitch type, release characteristics, location, and outcome). Batter decisions, such as whether to swing, and pitch results are represented explicitly as tokens. This representation allows the LLM to ingest heterogeneous information in a unified format.

\subsection{World Model Training}

We initialize our model from a pretrained Llama-family architecture~\cite{grattafiori2024llama3herdmodels} and continue pretraining on baseball event sequences using standard next-token prediction. Training is performed on regular-season data only, ensuring that postseason evaluation represents an out-of-distribution setting. Due to the length of games---averaging over 120K tokens per game---we split sequences into sliding windows of 3072 tokens without overlap.

\subsection{Downstream Prediction}

Prediction tasks such as pitch type classification and swing decision prediction are cast as conditional generation or constrained decoding problems. Given the preceding context up to a pitch, the model predicts the next event token, which is then mapped back to structured labels for evaluation.

\section{Data and Training}

\paragraph{Data collection.} We collect over ten years of MLB tracking data from Statcast and the MLB StatsAPI, yielding more than seven million pitch sequences and approximately three billion tokens. The dataset covers regular-season and postseason games; however, only regular-season data are used for training. This strict separation enables a clean evaluation of out-of-domain generalization.

The original data format is tabular. To align with LLM inputs, we directly convert the data into texts chronologically, including game details like release point, RPM, speed, plate coordinates, spin direction, swing speed; and also the player decisions (Figure~\ref{fig:data}).

\paragraph{Training details.} Training is conducted using continuous pretraining on the serialized sequences. All baseline methods are trained on the same temporal span of data for fair comparison, following standard preprocessing and filtering protocols described in prior work. We use Llama-3.2 3B~\cite{grattafiori2024llama3herdmodels} as the backbone model, considering it's compact size and cutting-edge performance. We train the model with batch size of 1024 and sequence length of 3072 on TPU-v4-64.

\section{Experimental Setup}

We evaluate our world model on two representative play-by-play prediction tasks that have been widely studied in prior baseball analytics literature:

\begin{itemize}
    \item \textbf{Pitch Type Prediction}: predicting the categorical type of the next pitch (\textit{e.g.}, four-seam fastball, slider, changeup) conditioned on the full pitch context.
    \item \textbf{Batter Swing Decision}: predicting whether the batter will attempt a swing given the pitch characteristics and game state.
\end{itemize}

Both tasks are formulated as conditional next-event prediction problems. For our language-model-based world model, predictions are obtained by autoregressively generating the next event token and mapping it to the corresponding structured label.

\paragraph{Baselines.}
We compare our approach against several strong neural baselines reproduced from prior work, each representing a different modeling paradigm. First, we include a recurrent neural network (RNN) baseline based on \citet{yifanpredict}, which uses an LSTM with pitcher and batter embeddings to perform binary fastball prediction. Second, we reproduce the attention-based LSTM model of \citet{yulstm}, which predicts multi-class pitch types using short pitch-history sequences augmented with contextual game features. Third, for batter swing decision prediction, we adopt the feed-forward multilayer perceptron (MLP) model proposed by \citet{gopal2024baseball}, which predicts pitch outcomes based on pitch physics, location, and batter statistics. All baseline models are trained and evaluated on the same temporal span of data as our world model to ensure a fair comparison.

\paragraph{Evaluation Metrics.}
Model performance is measured using accuracy, recall, and F1 score where applicable. For pitch type prediction, we primarily report accuracy and macro-averaged F1 to account for class imbalance among pitch types. For batter swing decision prediction, we additionally report in-zone and out-of-zone accuracy to better capture decision quality relative to pitch location. Together, these metrics provide a comprehensive view of both overall predictive performance and behavior under different contextual conditions.

\section{Results}

\paragraph{Prediction performance.} Our world model achieves competitive or state-of-the-art performance across tasks using a single LLM backbone. For pitch type prediction (Table~\ref{tab:pitch}), our model slightly improves over the RNN baseline of \citet{yifanpredict}, achieving an F1 score of 0.722 compared to 0.720, while matching recall at 0.792. For batter swing decision prediction, our model substantially outperforms the MLP baseline of \citet{gopal2024baseball}, achieving 0.766 in-zone accuracy and 0.792 out-of-zone accuracy, compared to 0.325 and 0.704 respectively (Table~\ref{tab:batting}).

Across tasks, the model correctly predicts approximately 84\% of next pitches within a plate appearance and 78\% of batter swing decisions when conditioned on full context. Notably, these results are obtained while evaluating on postseason data, which prior work typically treats as out-of-distribution.

\begin{table}[t]
\centering
\begin{tabular}{lccc}
\toprule
\textbf{Pitch Prediction} & \textbf{Accuracy} & \textbf{Recall} & \textbf{F1} \\
\midrule
\citet{yifanpredict} & 0.633 & \textbf{0.792} & 0.720 \\
Ours & \textbf{0.637} & \textbf{0.792} & \textbf{0.722} \\
\bottomrule
\end{tabular}
\caption{Pitch type prediction (Fastball vs Non-fastball) results comparing prior work and our model.}
\label{tab:pitch}
\end{table}

\begin{table}[t]
\centering
\begin{tabular}{lcc}
\toprule
\textbf{Batting Decision} & \textbf{IZ Acc} & \textbf{OZ Acc} \\
\midrule
\citet{gopal2024baseball}& 0.325 & 0.704 \\
Ours & \textbf{0.766} & \textbf{0.792} \\
\bottomrule
\end{tabular}
\caption{Batting decision accuracy compared to the prior work. IZ and OZ stand for in-zone pitch and out-zone pitch, where the accuracy is measured whether the model successfully predicted batter's swing decision.}
\label{tab:batting}
\end{table}

\begin{figure}
    \centering
    \includegraphics[width=\linewidth]{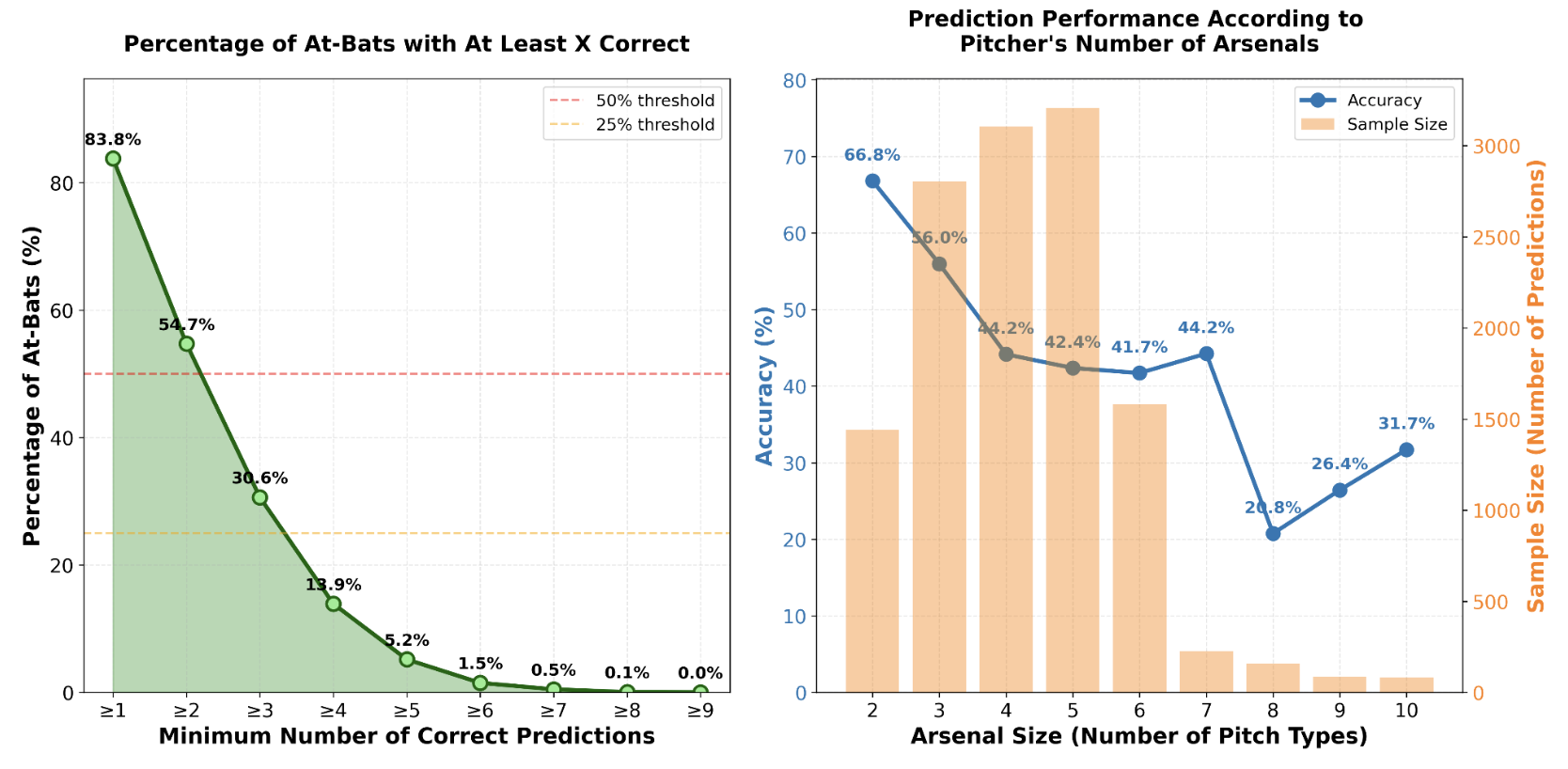}
    \caption{Left: prediction errors accumulate over longer sequences. Right: the increased uncertainty and strategic variability introduced by pitchers with more diverse repertoires.}
    \label{fig:bat_consistency}
\end{figure}

\paragraph{Consistency analysis.} Figure~\ref{fig:bat_consistency} (left) shows the percentage of plate appearances in which the model correctly predicts at least $X$ pitches. We observe that 83.8\% of at-bats contain at least one correct prediction, and 54.7\% contain at least two correct predictions, indicating that correct predictions often persist across multiple pitches within the same plate appearance rather than occurring in isolation. However, the proportion decreases rapidly as the required number of correct predictions increases, suggesting that prediction errors accumulate over longer sequences. This behavior highlights a fundamental challenge of auto-regressive world models when applied to extended play-by-play horizons.

Figure~\ref{fig:bat_consistency} (right) shows pitch type prediction accuracy as a function of a pitcher’s arsenal size, measured by the number of distinct pitch types. The model achieves its highest accuracy for pitchers with smaller arsenals (two to three pitch types), with accuracy reaching 66.8\%, while performance degrades substantially as arsenal size increases. This trend reflects the increased uncertainty and strategic variability introduced by pitchers with more diverse repertoires. In addition, the reduced number of samples for large-arsenal pitchers further compounds the difficulty of learning reliable prediction patterns in these regimes.

\begin{figure}
    \centering
    \includegraphics[width=\linewidth]{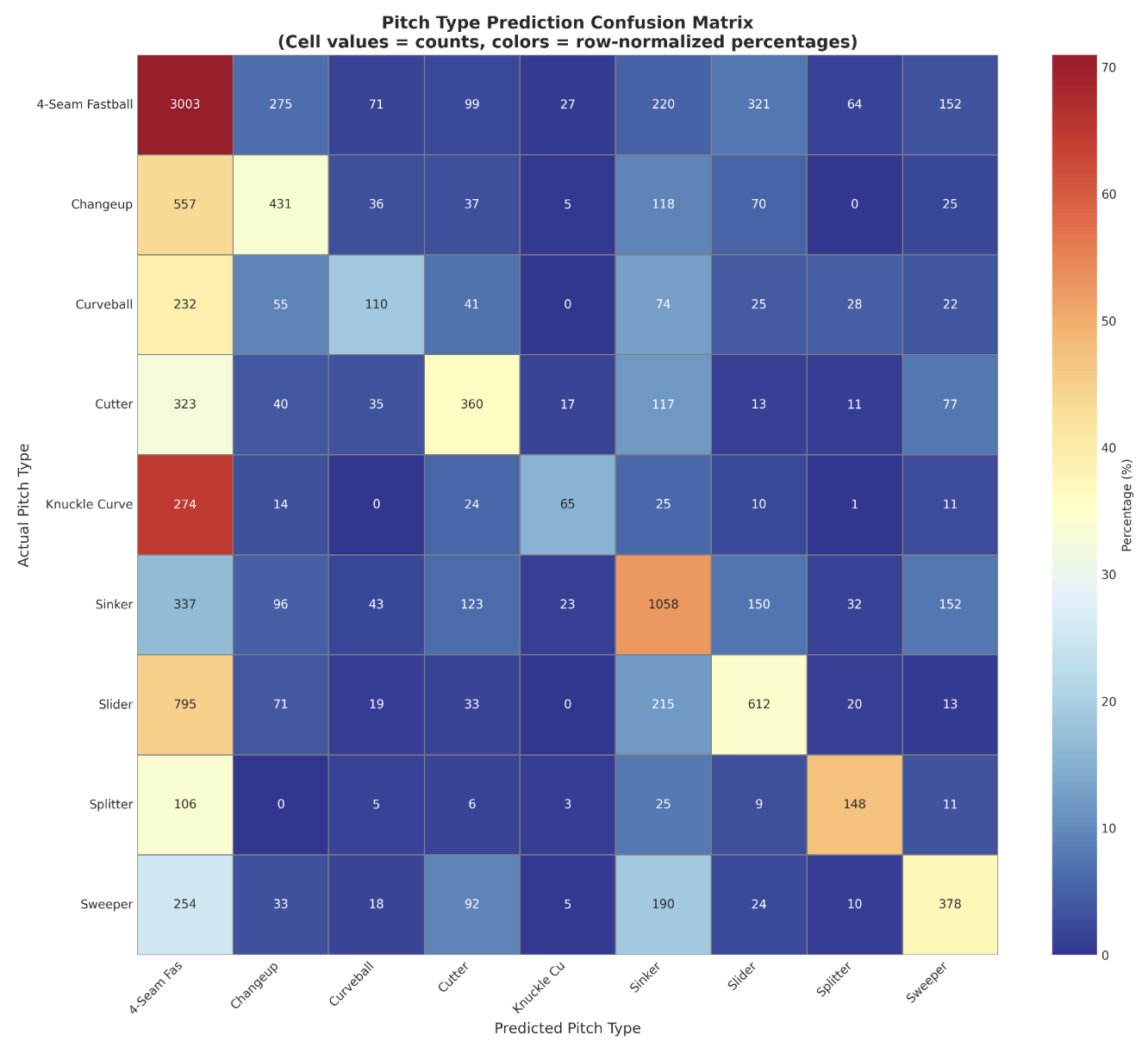}
    \caption{The model exhibits a strong bias toward predicting the most frequent pitch types.}
    \label{fig:confusion_matrix}
\end{figure}

\paragraph{Pitch type analysis.} Figure~\ref{fig:confusion_matrix} presents the confusion matrix for pitch type prediction. The model exhibits a strong bias toward predicting the four-seam fastball, which is both the most frequent pitch in the dataset and the most common misclassification target for other pitch types such as sliders, changeups, and sinkers. This indicates a frequency-driven prior under uncertainty. Furthermore, pitch types with similar physical and movement characteristics---for example, slider versus sweeper or sinker versus changeup---display elevated mutual confusion, consistent with known similarities in pitch design.

\begin{figure}
    \centering
    \includegraphics[width=\linewidth]{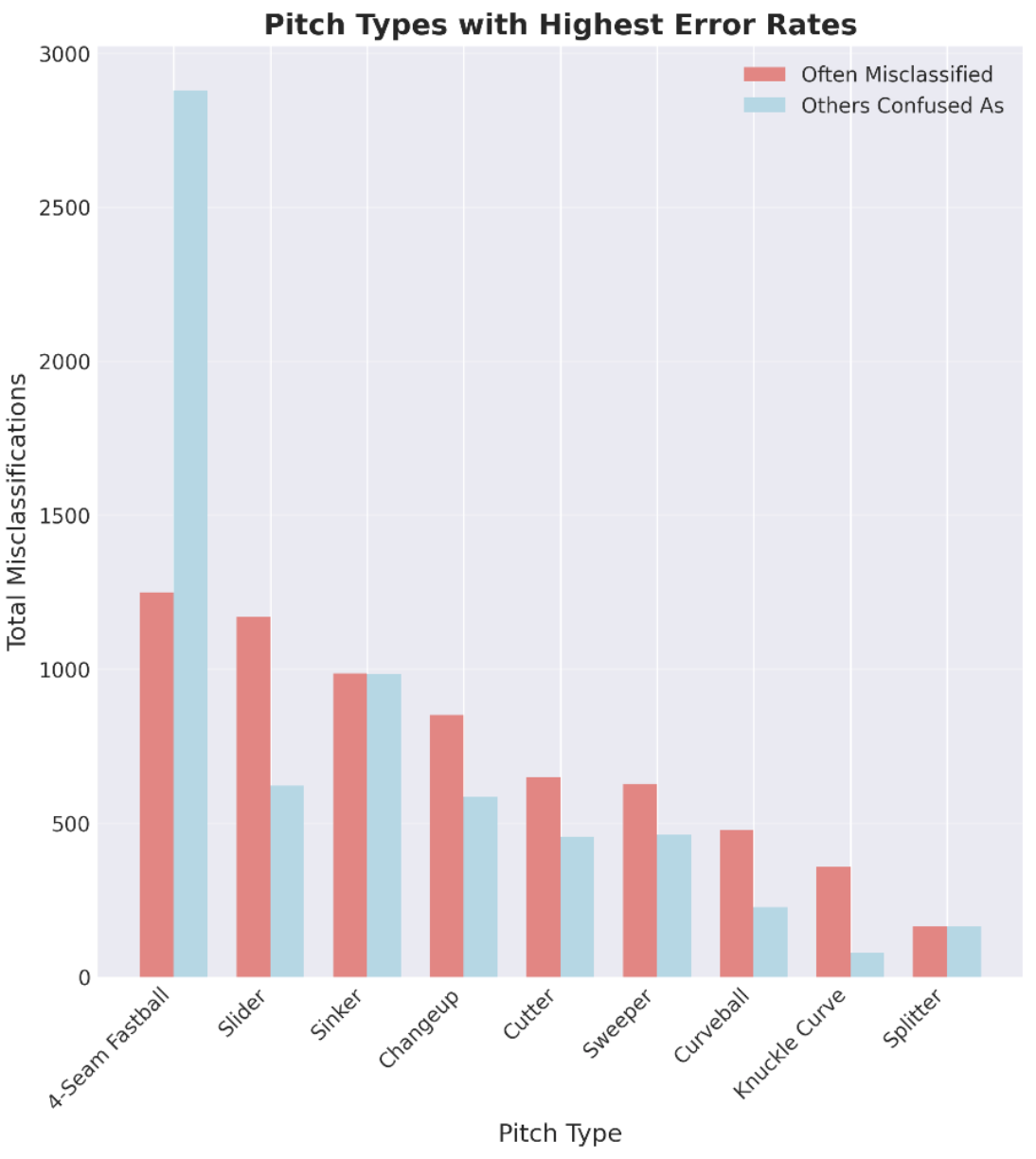}
    \caption{Four-seam fastballs and sliders dominate the error distribution, while rarer pitch types incur substantially fewer errors.}
    \label{fig:error_analysis}
\end{figure}

\paragraph{Error analysis.} Our error analysis (Figure~\ref{fig:error_analysis}) summarizes the absolute misclassification counts by pitch type. Four-seam fastballs and sliders dominate the error distribution, largely due to their high prevalence in the data rather than uniquely poor per-class performance. In contrast, rarer pitch types such as knuckle curves and splitters incur substantially fewer errors, reflecting both smaller sample sizes and more distinctive physical signatures. These results underscore the importance of interpreting prediction errors jointly with class frequency and contextual ambiguity.

\section{Discussion}

Qualitative and quantitative analyses reveal several important behaviors. First, the model exhibits a bias toward predicting four-seam fastballs, reflecting their prevalence in MLB data. This bias is consistent with prior neural baselines and highlights the need for richer player-specific conditioning. Future works could focus on retrieval-augmented player profiles.

Second, the model demonstrates strong robustness under distribution shift. By training exclusively on regular-season data and evaluating on postseason games, we show that a large-scale world model can generalize across strategic and contextual changes, such as altered pitch mixes and higher-leverage situations.

Finally, the LLM framework naturally supports extensibility. New data modalities, such as batting tracking features introduced in 2024, can be incorporated without redesigning task-specific architectures, reinforcing the value of a unified world model approach.

\section{Conclusion}

We present Neural Sabermetrics with World Model, a large language model trained to predict play-by-play dynamics in baseball. By treating baseball tracking data as a sequential language and leveraging continuous pretraining, our model unifies multiple prediction tasks within a single backbone and achieves strong performance, including robust generalization to postseason data. Our results suggest that world models offer a promising foundation for the next generation of predictive sports analytics.

\section*{Limitations}

Our work has several limitations. Due to context-length constraints, long games must be segmented into fixed-size windows, potentially breaking long-range dependencies. It remains an open question whether token-based representations or latent game-state abstractions are more effective for long-horizon world modeling. Additionally, while we demonstrate strong results on pitch and swing prediction, extending evaluation to richer outcomes such as batted-ball trajectories and run expectancy is an important direction for future work.

\section*{Acknowledgement}
We would like to thank Google’s TPU Research Cloud (TRC) for providing Cloud TPU resources. We also acknowledge the use of the Babel HPC Cluster at Carnegie Mellon University for the computational support provided for this work.

\bibliography{custom}

\appendix



\end{document}